\documentclass[10pt,twocolumn,letterpaper]{article}

\usepackage[pagenumbers]{iccv} %

\usepackage{times}
\usepackage{epsfig}
\usepackage{graphicx}
\usepackage{amsmath}
\usepackage{amssymb}
\usepackage{colortbl} %
\usepackage{enumitem}
\usepackage{wrapfig}
\usepackage{multirow}
\usepackage{cuted}
\usepackage{mathtools}
\usepackage{tikz}
\usepackage[english]{babel}
\usepackage{blindtext}
\usepackage{mathtools}
\usepackage{xspace}
\usepackage{booktabs}

\newcommand{\parsection}[1]{

\vspace{3pt}
\noindent\textbf{#1}\xspace }

\definecolor{tabfirst}{rgb}{1, 0.7, 0.7}
\definecolor{tabsecond}{rgb}{1, 0.85, 0.7}
\definecolor{tabthird}{rgb}{1, 1, 0.7}

\newif\ifshowtodos
\showtodostrue %

\newcommand{\methodname}{FlowR}

\newcommand{\fs}{\cellcolor{tabfirst}}
\newcommand{\nd}{\cellcolor{tabsecond}}
\newcommand{\rd}{\cellcolor{tabthird}}

\newcommand{\loss}{\mathcal{L}}
\newcommand{\real}{\mathbb{R}}
\newcommand{\Exp}{\mathbb{E}}
\newcommand{\normal}{\mathcal{N}}
\newcommand{\identity}{\mathbf{I}}
\newcommand{\zero}{\mathbf{0}}

\newcommand{\diag}[1]{\operatorname{diag}(#1)}

\newcommand{\pose}{\mathbf{P}}
\newcommand{\rot}{\mathbf{R}}
\newcommand{\grot}{\mathbf{U}}
\newcommand{\trans}{\mathbf{t}}
\newcommand{\intr}{\mathbf{K}}
\newcommand{\ims}{\mathcal{I}}
\newcommand{\im}{\mathtt{I}}
\newcommand{\depth}{\mathtt{D}}
\newcommand{\viewgraph}{G_\text{vis}}
\newcommand{\nodes}{\mathcal{I}}
\newcommand{\edges}{\mathcal{E}}

\newcommand{\colvec}{\mathbf{c}}
\newcommand{\loc}{\mathbf{p}}
\newcommand{\pix}{\mathbf{p}'}
\newcommand{\Gauss}{\mathcal{G}}
\newcommand{\gauss}{\mathfrak g}
\newcommand{\mean}{\boldsymbol\mu}
\newcommand{\cov}{\mathtt{\Sigma}}
\newcommand{\opa}{\alpha}
\newcommand{\quat}{\mathbf{q}}
\newcommand{\svec}{\mathbf{s}}

\newcommand{\velocity}{\boldsymbol v}

\newcommand{\latent}{\mathbf{z}}
\newcommand{\cond}{\mathbf{y}}

\definecolor{iccvblue}{rgb}{0.21,0.49,0.74}
\usepackage[pagebackref,breaklinks,colorlinks,allcolors=iccvblue]{hyperref}
\usepackage[capitalize]{cleveref}

\def\paperID{****}
\def\confName{ICCV}
\def\confYear{2025}

\title{
\vspace{-15pt}
\methodname: Flowing from Sparse to Dense 3D Reconstructions}

\author{Tobias Fischer$^{1,2}$ \qquad
 Samuel Rota Bul\`{o}$^{2}$ \qquad
 Yung-Hsu Yang$^{1}$ \qquad
 Nikhil Keetha$^{2,3}$ \qquad \\
 Lorenzo Porzi$^{2}$ \qquad
 Norman Müller$^{2}$ \qquad
 Katja Schwarz$^{2}$ \qquad
 Jonathon Luiten$^{2}$ \qquad \\
 Marc Pollefeys$^{1}$ \qquad 
 Peter Kontschieder$^{2}$ \vspace{0.05cm}\\
{ $^{1}$ ETH Z{\"u}rich \quad $^{2}$ Meta Reality Labs Zürich \quad $^{3}$ CMU}\vspace{0.05cm} \\
   \small{\url{https://tobiasfshr.github.io/pub/flowr} }
}

\newread\imgstream
\immediate\openin\imgstream=imagedata.in
\makeatletter
\def\new@kvginclip#1{}
\def\new@kvgintrim#1{}
\let\old@kvginclip\KV@Gin@clip
\let\old@kvgintrim\KV@Gin@trim
\let\oldincludegraphics\includegraphics
\providecommand{\includegraphics}{}
\renewcommand{\includegraphics}[2][]{%
  \immediate\read\imgstream to \src
  \immediate\read\imgstream to \removecrop
  \ifnum\removecrop=1
      \let\KV@Gin@clip\new@kvginclip
      \let\KV@Gin@trim\new@kvgintrim
  \fi
  \oldincludegraphics[#1]{\src}%
  \let\KV@Gin@clip\old@kvginclip
  \let\KV@Gin@trim\old@kvgintrim}
\makeatother

\begin{document}
\maketitle

\begin{abstract}
3D Gaussian splatting enables high-quality novel view synthesis (NVS) at real-time frame rates. However, its quality drops sharply as we depart from the training views. Thus, dense captures are needed to match the high-quality expectations of applications like Virtual Reality (VR). 
However, such dense captures are very laborious and expensive to obtain.
Existing works have explored using 2D generative models to alleviate this requirement by distillation or generating additional training views.
These models typically rely on a noise-to-data generative process conditioned only on a handful of reference input views, leading to hallucinations, inconsistent generation results, and subsequent reconstruction artifacts.
Instead, we propose a multi-view, flow matching model that learns a flow to directly connect novel view renderings from possibly sparse reconstructions to renderings that we expect from dense reconstructions. This enables augmenting scene captures with consistent, generated views to improve reconstruction quality.
Our model is trained on a novel dataset of 3.6M image pairs and can process up to 45 views at $540\times960$ resolution (91K tokens) on one H100 GPU in a single forward pass. Our pipeline consistently improves NVS in sparse- and dense-view scenarios, leading to higher-quality reconstructions than prior works across multiple, widely-used NVS benchmarks.

\end{abstract}

\section{Introduction}

\begin{figure}[t!]
    \centering
    \includegraphics[width=\linewidth]{figures/teaser.pdf}\vspace{-2mm}
    \caption{\label{fig:teaser}\textbf{Flowing from sparse to dense 3D reconstructions.} Contrary to previous diffusion and flow matching models that map a standard multivariate Gaussian distribution $p_0(\latent)$ to an, often conditional, target distribution $p_1(\latent \mid \mathbf{y})$, we consider \emph{source} distributions of the form $p_0(\latent \mid \mathbf{y})$. We use novel view renderings of sparse reconstructions as source distribution samples, which we map to the target distribution $p_1(\latent \mid \mathbf{y})$ that represents reconstructions obtained under optimal, dense conditions (\ie ground truth).}\vspace{-1mm}
\end{figure}

3D reconstruction is the process of estimating the 3D scene geometry and appearance from a set of 2D images. Given a large and dense enough set of images, modern 3D reconstruction methods such as neural radiance fields (NeRF)~\cite{mildenhall2021nerf} and 3D Gaussian splatting (3DGS)~\cite{kerbl20233d} produce 3D representations that can be rendered into novel views that are almost indistinguishable from reality. This is the task of novel view synthesis (NVS), and it enables various applications such as immersive VR experiences inside captured real-world scenes. However, for these methods to achieve such a high level of photorealism, they need an extremely large number of captured images per scene. Obtaining these images is laborious and not always possible. Thus, one of the core challenges in 3D reconstruction and novel view synthesis is how to build an algorithm that can achieve equally good results with far fewer images, while also being able to take advantage of a large set of images when they are available. 

In this paper, we tackle this problem with our \methodname{} method. \methodname{} consists of two parts: (a) a robust initial reconstruction pipeline based on 3DGS~\cite{kerbl20233d} but designed for both sparse- and dense-view settings; and (b) a data densification procedure, which uses flow matching~\cite{liu2023flow} to generate high-quality extra views that can be used to improve reconstruction.

Several previous approaches~\cite{poole2023dreamfusion, wang2023prolificdreamer, wu2024reconfusion, gao2024cat3d, yu2024viewcrafter, chen2024mvsplat360, liu20253dgs} have used generative models to generate novel views to improve reconstruction.
These works rely on diffusion~\cite{ho2020denoising} or flow matching~\cite{lipman2022flow} models that map samples between a data and a noise distribution. As such, these works use a noise-to-data generative process, usually conditioned on initial input images or renderings (\cref{fig:teaser}, top).
We hypothesize that a generative process that directly maps samples from a conditional source distribution of rendered images to the target distribution of ground truth images is preferable to conditional noise-to-data generation.
Additionally, we observe that flow matching is a paradigm for generative modeling where the model learns a velocity field that transports samples from \emph{any} source to a target distribution.
Therefore, in this work, we pose the problem as a flow matching problem where, instead of modeling the velocity field between noise and data, we model the velocity field between the distribution of incorrect novel view renderings and the respective real images of that viewpoint (\cref{fig:teaser}, bottom). 

In this way, if we already have enough dense input images that our initial reconstruction is good enough for a particular view, the flow matching model can simply learn not to change the input. This formulation ensures that the generative model does not hallucinate unnecessary new details that are inconsistent and conflict with existing scene content, and thus results in sharp scene details, avoiding blurred-out averages of inconsistent generations.

Our flow matching model generates $N$ images simultaneously with a single multi-view diffusion transformer to ensure that all generated images are consistent with each other, while also conditioning the generation of these $N$ images with $M$ input images from the initial reconstruction to ensure that the novel generated views are consistent with the input views.
To train our model, we create a dataset of 10.3k reconstructed scenes using our robust initial reconstruction approach, from which we obtain 3.6M pairs of novel view reconstructions with their corresponding ground truth images.
In summary, we introduce \methodname, a pipeline that bridges the gap between sparse and dense 3D reconstructions. Our contributions are as follows.

\begin{itemize}
    \item We develop a fast initial 3D reconstruction pipeline that produces 3DGS representations from arbitrary view distributions and use it to collect a large-scale dataset of 3.6M rendered and ground-truth image pairs.
    \item We propose a flow matching formulation that directly incorporates novel-view renderings from initial reconstructions as a surrogate for the initial ``noise" distribution and use it to train a multi-view flow matching model that enhances novel-view renderings.
    \item We demonstrate that our trained model enhances the quality of 3D reconstructions in both sparse and dense view scenarios by simply incorporating novel-view renderings enhanced by our flow matching model into the reconstruction process.
\end{itemize}
Our approach outperforms prior methods in both sparse- and dense-view scenarios across three NVS benchmarks.

\section{Preliminaries}
\label{sec:preliminaries}
We describe the core ideas that our approach is built upon, specifically 3D Gaussian splatting (\cref{sec:3dgs}) and flow matching (\cref{sec:fm}).

\subsection{3D Gaussian Splatting}
3D Gaussian splatting~\cite{kerbl20233d} describes a method for NVS from a set of calibrated images.
\label{sec:3dgs}
\parsection{Representation.}
Let $\ims \coloneqq \{\ims_i\coloneqq(\im_i, \pose_i,\intr_i)\}_{i=1}^N$ denote a set of $N$ input images $\im_i$ with camera-to-world transformation $\pose_i \coloneqq [\rot_i|\trans_i]$ and intrinisc matrix $\intr_i$. We are interested in representing the underlying 3D scene as a set of 3D Gaussian primitives $\Gauss\coloneqq\{\mathfrak g_k\}_{k=1}^K$. Each 3D Gaussian $\gauss_k \in \Gauss$  is parametrized by $\gauss_k \coloneqq \{ \mean_k, \svec_k, \quat_k, \opa_k, \colvec_k\}$, \ie position, scale, rotation, opacity and view-dependent color, respectively. The Gaussian kernel takes the following form
\begin{equation}
    \mathfrak g_k(\loc) \coloneqq \operatorname{exp} \left( -\frac{1}{2} [\loc - \mean_k]^\top \cov_k^{-1} [\loc - \mean_k] \right)\,.
\end{equation}
Here, the covariance matrix factorizes as $\cov_k \coloneqq \grot_k \diag{\svec_k}^2\grot_k^\top$, where $\grot_k$ is the rotation matrix corresponding to quaternion $\quat_k$ and $\svec_k \in \real_{+}^{3}$ entails positive scaling factors turned into a diagonal matrix with $\diag{}$.

\parsection{Rendering.}
To render the 3D scene from a camera $c$, we map the 3D Gaussians to the image plane. 
In particular, for each primitive $\gauss_k$, let $\gauss_k^c$ denote a 2D Gaussian kernel with its mean $\mean_k^c$ defined as the primitive's position projected to the image plane, \ie $\mean_k^c\coloneqq\Pi^c(\mean_k)$, and its covariance defined as $\cov_k^c\coloneqq\mathtt J_k^c\cov_k\mathtt J_k^{c\top}$, where $\mathtt J_k^c$ is the Jacobian of the 3D-to-2D projection function $\Pi^c$ evaluated at $\mean_k$. To render the color of pixels $\pix$ of camera $c$, we apply alpha compositing on the depth-sorted primitives $\gauss_k$:
\begin{equation}\label{eq:rendering}
    \colvec(\pix) \coloneqq \sum_{k=0}^{K} \colvec'_k w_k\prod_{j=0}^{k-1}(1 - w_j)\,,
\end{equation}
where $\quad w_k \coloneqq \alpha_k \mathfrak g^c_k(\pix)$ and $\colvec'_k$ is the evaluated spherical harmonics function of $\colvec_k$ at the view direction of $c$.

\parsection{Optimization.}
Using the differentiable rasterizer in~\cite{kerbl20233d}, we fit the set of 3D Gaussians $\Gauss$ to the training images $\ims$ by applying the following per-image loss function:
\begin{equation}
\loss_\text{GS}(\im;\mathcal G) \coloneqq (1-\lambda_\text{ssim}) \|\hat{\im} - \im \|_1  + \lambda_\text{ssim}\operatorname{SSIM}(\hat{\im}, \im)\,,
\label{eq:loss_3dgs}
\end{equation}
where $\hat{\im}$ is a rendering as per~\cref{eq:rendering} from the viewpoint of training image $\im$, $\lambda_\text{ssim} \coloneqq 0.2$, and $\operatorname{SSIM}(\cdot)$ is the structural similarity loss~\cite{wang2004image}. The optimization is interleaved with an adaptive density control (ADC) mechanism~\cite{kerbl20233d} that prunes transparent primitives, and splits and clones primitives in over- and under-reconstructed areas, respectively.

\begin{figure*}[t!]
    \centering
    \includegraphics[width=\textwidth]{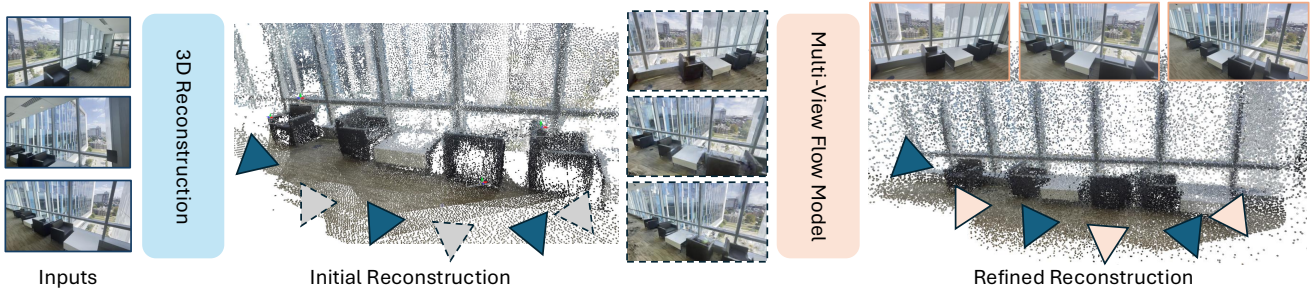}\vspace{-3mm}
    \caption{\label{fig:overview}\textbf{Overview.} Given a set of source input images $\ims_\text{src}$ (blue), we use our robust reconstruction method (\Cref{sec:reconstruct_generate}) to create an initial reconstruction result that can be rendered at various perspectives (gray). We use these renderings as the source samples of our flow matching model (\Cref{sec:method_model}), which maps the rendered images to the target distribution, \ie ground-truth images. We use the generated views (orange) to improve the quality of the reconstruction (\Cref{sec:generate_reconstruct}).}\vspace{-2mm}
\end{figure*}
\subsection{Flow Matching}\label{sec:fm}
Flow matching is a paradigm for generative modeling that we briefly introduce in this section and refer to the original papers~\cite{lipman2022flow,liu2023flow} for further details.

\parsection{Model.} Let $\latent_0$ denote a sample from a source distribution $p_0$ that we wish to map to a sample $\latent_1$ of a target distribution $p_1$.  
We model this transformation as a continuous probability flow with time $t \in [0, 1]$, governed by an ordinary differential equation (ODE):
\begin{equation} \label{eq:ode}
d\latent_t = \velocity(\latent_t, t)dt
\end{equation}
where $\velocity(\latent_t, t)$ is a time-dependent velocity field that induces a probability path $p_t$ interpolating $p_0$ and $p_1$. Assuming an optimal transport path, the state $\latent_t$ should satisfy
\begin{equation}
    \latent_t \coloneqq t \latent_1 + (1-(1-\sigma_\text{min})t)\latent_0
\label{eq:flow_state}
\end{equation}
where $\sigma_\text{min} \coloneqq 10^{-5}$. It follows from differentiating the equation w.r.t. $t$ that the true velocity field is given by
\begin{equation}\label{eq:velocity}
    \velocity_t \coloneqq \frac{d\latent_t}{dt} = \latent_1 - (1-\sigma_\text{min})\latent_0\,.
\end{equation}
\parsection{Training.}
We train a network with parameters $\theta$ to approximate $\velocity_t$ with the conditional flow matching loss~\cite{lipman2022flow, albergo2022building}:
\begin{equation}
   \loss_\text{CFM}(\theta) \coloneqq \Exp_{t,\latent_0,\latent_1} \| \velocity_\theta(\latent_t, t) - \velocity_t \|_2^2
\label{eq:loss_cfm}
\end{equation}
where $\latent_t$ and $\velocity_t$ are given as per~\cref{eq:flow_state,eq:velocity} with $\latent_0$ and $\latent_1$ sampled from $p_0$ and $p_1$, respectively, and $t$ sampled from a logit normal distribution~\cite{esser2024scaling}.
Note that, although this formulation can transport samples between two arbitrary distributions, in practice $p_0$ is often a standard Gaussian distribution $\normal(\zero, \identity)$.

\parsection{Inference.}
To transport a sample $\latent_0$ from $p_0$ to $p_1$, we numerically solve the ODE in~\cref{eq:ode} using the estimated velocity $\velocity_\theta(\latent_t, t)$ and Euler's method, but other numerical solvers are viable.

\section{Method}
Our goal is to improve the quality of an initial reconstruction of a scene created from a possibly sparse set of posed images. We exploit flow matching to improve the quality of novel view renderings from the initial reconstruction, which serve as auxiliary training data to fit an improved 3D representation.
Although our method could work with different scene representations, it is convenient to adopt Gaussian primitives because they can be trained fast. 
In fact, since our approach is data-driven, we need to construct a dataset of reconstructions under different sparsity levels, and Gaussian splatting ensures better scalability for this purpose.
In \cref{sec:reconstruct_generate}, we describe our robust 3D reconstruction method that serves as a basis for generating our dataset of 3D reconstructions and provides the initial scene reconstructions. %
\cref{sec:method_model} describes how we employ flow matching to improve novel-view renderings from the initial reconstructions, which, in turn, are used to improve our reconstruction results in \cref{sec:generate_reconstruct}. See \cref{fig:overview} for an illustration.

\subsection{Robust 3D Reconstruction}
\label{sec:reconstruct_generate}

We aim to devise a 3D reconstruction pipeline that produces semi-dense, metric-scale point clouds from arbitrary input view distributions, \ie ranging from a few sparse to thousands of dense views. To this end, we combine recent advances in learning-based structure-from-motion (SfM) and monocular depth estimation with classical SfM tools.

First, we define a sparse co-visibility graph $\viewgraph \coloneqq (\nodes, \edges)$ on the set of posed input images $\ims$ similar to~\cite{duisterhof2024mast3rsfm}. We use keyframing to limit the number of view pairs for matching and thus enable a scalable 3D reconstruction process. In particular, we sample $\sqrt{N}$ keyframes using farthest point sampling. The edges $\edges\subset\ims\times\ims$ are constructed by densely connecting all keyframes and connecting other nodes to their closest keyframe and their $k$ nearest neighbors. For both keyframes selection and edge construction we consider the simple image co-visibility metric, $\mathcal S(\ims_i,\ims_j)\coloneqq 1- \frac{\mathcal D(\ims_i, \ims_j)}{\max_{ij}\mathcal D(\ims_i, \ims_j)}$, where 
\begin{equation}\label{eq:viewdist}
\mathcal D(\ims_i, \ims_j) \coloneqq \|\rot_i - \rot_j\|_F + \eta \| \trans_i - \trans_j \|_2\,.
\end{equation}
Here, $\| \cdot \|_F$ is the Frobenius norm and $\eta\coloneqq \frac{1}{6}$.

Second, we run MASt3R~\cite{leroy2025grounding} on all edges $\edges$ to obtain 3D points and descriptors for each view pair.
We extract matches following~\cite{leroy2025grounding} with a coarse-to-fine nearest-neighbor search on the descriptors. Since extracting dense matches is useful for dense geometry reconstruction from a small number of observations, but not beneficial for scenes with many images, we subsample matches using confidence and pixel tolerance thresholds based on the number of available input views.

Third, given the feature matches for each pair of views in $\edges$, we can take advantage of $\viewgraph$ to create feature tracks by exploiting the transitivity of the matches. 
Finally, we use COLMAP~\cite{schoenberger2016sfm} to triangulate 3D points from the given feature tracks. 
To achieve a 3D reconstruction on a metric scale, we align the reconstructed depth $\depth$ with the monocular depth estimates $\hat{\depth}$ of~\cite{hu2024metric3d} by computing a scalar scaling factor $\beta$ solving $\min_{\beta} \sum_{i=1}^N \mathtt{W}_i  \|\beta \depth_i - \hat{\depth}_i \|_1$, where $\mathtt{W}_i$ is the confidence of the monocular depth estimate $\hat{\depth}_i$.

\parsection{Fitting 3D Gaussians.}
Using this reconstruction as an initialization, we fit 3D Gaussians $\Gauss$ from the input images with a short schedule. We train for $5$k steps with a warm-up phase of 200 steps. After warm-up, we use ADC with AbsGrad~\cite{ye2024absgs} metric until step $2.5$k.
Compared to previous works~\cite{fan2024instantsplat} that use 3D points obtained from global pointcloud alignment~\cite{leroy2025grounding} as initialization, our pipeline has two advantages. First, it refines the geometry by using track constraints. Second, it removes duplicated and inconsistent 3D points from the initialization, improving performance and scalability (see \cref{sec:ablations}).

\parsection{Dataset generation.}
We leverage our robust reconstruction pipeline and two recent large-scale NVS benchmarks~\cite{ling2024dl3dv, yeshwanth2023scannet++} to generate a dataset of rendered and ground-truth image pairs. For each scene, we first select an input view subset with a varying degree of view sparsity to achieve a large variety of view distributions (see \cref{sec:app_data}). Next, we utilize all views not in this subset as privileged information for our flow matching model. We reconstruct the scenes from the input view subset with our aforementioned pipeline and create rendering, ground-truth pairs from all other available views, producing a training dataset of $3.6$M image pairs from $10.3$k sequences. Processing all scenes took about $1.1$k GPU hours or 6.5 minutes on average per scene on budget GPUs like RTX 2080 Ti.

\subsection{Flowing from Sparse to Dense Reconstructions}
\label{sec:method_model}
We use flow matching to model a joint distribution of novel-view (target) images $\latent\coloneqq\left\{\im^\text{tgt}_i\right\}_i$ conditioned on $\cond\coloneqq(\ims_\text{src}, \mathcal P_\text{tgt})$, namely a set of (source) posed, reference images $\ims_\text{src}$ %
and a set of target novel-view poses $\mathcal P_\text{tgt}$. Ideally, it is supposed to match the true data distribution $p_1(\latent|\cond)$.
A classical way to implicitly fit $p_1(\latent|\cond)$ in the flow matching framework consists of learning a vector field $\velocity_\theta(\latent,\cond, t)$ that depends also on the conditioning variables $\cond$, which generates a (conditional) probability density path connecting a standard multivariate Gaussian as the source distribution $p_0(\latent)$ to the target distribution $p_1(\latent|\cond)$. 

Inspired by~\cite{liu2024flowing}, in our work we adopt a different strategy and condition also the source distribution $p_0$ on $\cond$, \ie we consider source distributions of the form $p_0(\latent|\cond)$. Now, fitting a 3D Gaussian representation $\mathcal G_\text{src}$ to the set of posed reference images $\ims_\text{src}$ can be interpreted as drawing a 3D representation from a stochastic process~\cite{kheradmand20253d}. Hence, rendering novel views at the poses $\mathcal P_\text{tgt}$ using $\mathcal G_\text{src}$ can be interpreted as drawing a sample from a conditional distribution $\latent|\cond$. Consequently, this process can characterize our conditional source distribution $p_0(\latent|\cond)$. By doing so, our model learns a flow connecting renderings $\latent_0\sim p_0(\latent|\cond)$ obtained from potentially sparse reconstructions (\eg if $\ims_\text{src}$ has few images) to renderings $\latent_1\sim p_1(\latent|\cond)$ that we would have under denser reconstructions (or even ground-truth) as enforced by our training procedure. 

Below, we detail the implementation of our multi-view flow model and in particular the velocity field $\velocity_\theta(\latent, \cond, t)$.

\parsection{Image encoding.}
In our model $\latent$ does not entail images in pixel space, but we use a pre-trained VAE encoder~\cite{esser2021taming} to encode each image to a latent tensor with dimensions $h\times w \times 16$. Similarly, conditioning images in $\cond_\text{src}$ undergo the same encoding in latent space before entering our model.
Finally, a pre-trained VAE decoder is used to map this latent representation back in pixel space.

\parsection{Camera conditioning.}
The network implementing the velocity field described below is conditioned on both source and target camera poses. We encode camera information as Plücker coordinate ray maps~\cite{zhang2024cameras, plucker1828analytisch} expressed in the coordinate frame of the reference view that is closest to the center of mass of all involved camera positions.

\parsection{Velocity field $\velocity_\theta(\latent, \cond, t)$.}
We input the target and source views as a single set of tokens to the flow matching model implemented as a diffusion transformer (DiT)~\cite{esser2024scaling, peebles2023scalable}.
All image latents provided as input, namely $\latent$ and the encoded source images from $\cond_\text{src}$, are first split into $2\times2$ patches. We project each patch to the token dimension and add a 2D positional encoding.
In addition, we encode the index $i$ of the image from which each token originates. To this end, we adopt a 1D sinusoidal encoding $\gamma(i)$~\cite{vaswani2017attention}. During training, the view index is chosen randomly from $[0, 1000]$, whereas at inference the view indices are $[0, N]$ where $N$ the number of images for a single forward pass.
Originally from LLM literature~\cite{chen2023extending}, this technique enables inference with very long context lengths and has recently been applied to 3D reconstruction~\cite{yang2025fast3r}.
The sequence of tokens is then processed with a series of transformer blocks to predict the output velocity tensor.
To share information across multiple reference and target views while leveraging large-scale text-to-image pre-training, we adapt the DiT block as follows.
We first collapse the view dimension into the batch dimension, keeping the self-attention layer equivalent to the pre-trained image model. Then, we collapse the view dimension into the spatial dimension.
We add the image index encoding $\gamma(i)$ to the state, concatenate it with the ray map embeddings, and linearly project the concatenated tokens back to the original dimension. We then insert a second self-attention layer to enable full \emph{multi-view attention}. Finally, before adding the output to the prior state of the block, we insert a zero linear layer~\cite{zhang2023adding} to keep the pre-trained initialization intact. See \cref{sec:app_method} for an illustration.

\subsection{Improving the 3D Reconstruction}
\label{sec:generate_reconstruct}

We sketch how we employ the flow matching model presented in the previous section to improve the \emph{initial} 3D reconstruction $\mathcal G_\text{src}$ that we obtain by running the procedure described in~\cref{sec:reconstruct_generate} on the initial source views $\ims_\text{src}$.  

\parsection{Generation of target views $\mathcal P_\text{tgt}$.}
We first select a set of camera poses suitable for augmenting the initial input views.
We distinguish between scenes where sparse input views lie on a continuous trajectory, \ie input views are ordered, and unordered photo collections. If input views are ordered, we model a smooth trajectory using B-spline basis functions from a sparse set of control points (source camera poses) and sample the target poses from it.
If not, we use the distance metric in \cref{eq:viewdist} to select reference poses with farthest point sampling. For each reference pose, we generate candidate points on a sphere with a random radius using Fibonacci sphere sampling~\cite{stollnitz1996wavelets}. We select the point with the maximum distance to the source cameras and apply a small perturbation to the reference view orientation to obtain the target camera pose. Finally, we filter target camera poses with too many points close to the camera or too few points inside the view frustum using the initial 3D reconstruction results, yielding the final set of target poses $\mathcal P_\text{tgt}$. 

\parsection{Renderings of target views.}
Renderings obtained from our initial reconstruction $\mathcal G_\text{src}$ at the target poses $\mathcal P_\text{tgt}$ can be considered a sample $\latent_0$ of the conditional prior distribution $p_0(\latent|\cond)$ as argued in~\cref{sec:method_model}.
Thus, the inference process of our flow matching model yields refined renderings $\latent_1$ starting from $\latent_0$. Once decoded, these renderings together with the respective poses $\mathcal P_\text{tgt}$ form additional training data denoted by $\mathcal I_\text{tgt}$ that complements the original set $\mathcal I_\text{src}$.

\parsection{Refined reconstruction $\mathcal G_\text{ref}$.}
We fit an improved 3D scene representation $\Gauss_\text{ref}$ on the union $\ims_\text{src}\cup\ims_\text{tgt}$ of source and target novel-view posed images. 
However, the two sets of images undergo different loss functions because views in $\ims_\text{tgt}$ could exhibit small imperfections, \eg, due to VAE compression artifacts.
Namely, we apply \cref{eq:loss_3dgs} to images in $\ims_\text{src}$ and
the following loss function to images in $\ims_\text{tgt}$:
\begin{align}
    \loss_\text{tgt}(\im; \Gauss) \coloneqq
        & (1 - \lambda'_\text{ssim}) \|\hat{\im} - \im \|_2  \notag \\
        &+ \lambda'_\text{ssim}\operatorname{SSIM}(\hat{\im}, \im)  \notag  + \lambda_\text{lpips}\operatorname{LPIPS}(\hat{\im}, \im)\,,
\end{align}
where $\lambda_\text{lpips} = \lambda'_\text{ssim} \coloneqq 0.02$ and $\operatorname{LPIPS}(\cdot)$ denotes perceptual similarity~\cite{zhang2018unreasonable}. Since color values are bound to $[0, 1]$, the L2 loss is less strict in terms of color variations. For the reconstruction, we employ the method provided in~\cref{sec:reconstruct_generate}, but restrict the selection of keyframes to $\ims_\text{src}$.
\section{Experiments}
\label{sec:experiments}

\parsection{Implementation details.}
We initialize our flow matching model from an image generation model trained on a large collection of image-text pairs similar to~\cite{esser2024scaling}. The base model has 2.7B parameters, and we drop the text conditioning layers, leaving us with 1.6B pre-initialized parameters. We train all multi-view layers from scratch; the total parameters are 1.75B. We use FlashAttention2~\cite{dao2023flashattention} for more efficient attention computation.
We train our model on 64 H100 GPUs for 125k steps (48 hours) at batch size 64 with 12 views per batch element and 512px width, keeping the original aspect ratio of each image. We use a cosine learning rate (LR) schedule with a warmup phase of 1k steps and a maximum LR of $10^{-5}$ scaled by $\sqrt{n}$ where $n$ is the number of batch elements.
We fine-tune our model at 960px width for 55k steps with 6 views per batch element for another 31 hours. We lower the maximum LR to $5 \cdot 10^{-6}$, keeping the same schedule.
For 3D Gaussian color $\colvec_k$, we use spherical harmonics degree of three. 

\parsection{Experimental setup.}
We compare to prior art across multiple benchmarks, \ie DL3DV140~\cite{ling2024dl3dv}, ScanNet$++$~\cite{yeshwanth2023scannet++}, and Nerfbusters~\cite{warburg2023nerfbusters}.
For DL3DV140, we first split each sequence into training and evaluation views and create two training splits by sampling $k \in \{12, 24\}$ equally spaced views from the training trajectory. We evaluate each sequence at the original $540 \times 960$ resolution. For ScanNet$++$, we use the validation set, follow the official training and testing splits, and evaluate at $640\times960$ resolution. For Nerfbusters, we follow the original evaluation protocol in~\cite{warburg2023nerfbusters}.
We measure view quality in terms of both low-level similarity via PSNR and SSIM~\cite{wang2004image}, and high-level, perceptual similarity using LPIPS~\cite{zhang2018unreasonable}.

\parsection{Baselines.}
We compare to state-of-the-art approaches that are open-source, \ie ViewCrafter~\cite{yu2024viewcrafter} and InstantSplat~\cite{fan2024instantsplat} in sparse-view settings, and GANeRF~\cite{roessle2023ganerf} in the dense-view setting. We also compare to Nerfbusters~\cite{warburg2023nerfbusters} on their proposed benchmark.

\subsection{Comparison to State-of-the-Art}
In \cref{tab:dl3dv140}, we compare to state-of-the-art sparse-view methods on the DL3DV~\cite{ling2024dl3dv} benchmark. In particular, we compare to InstantSplat~\cite{fan2024instantsplat} and ViewCrafter~\cite{yu2024viewcrafter}. For a fair comparison, we use the same prior information (\ie posed images) and downstream pipelines (\ie our refined reconstruction) if applicable.
While InstantSplat performs competitively to our \emph{initial} reconstructions in the 12-view setting, its performance stagnates in the 24-view setting, highlighting its limitation to sparse view scenarios. The generated views from ViewCrafter do not improve reconstructions meaningfully in our experiments, as its results with our downstream pipeline lag behind our initial reconstructions.
Our full method on the other hand improves 3D reconstruction results by a significant margin across all metrics, both in the 12-view and 24-view settings. The improvement is particularly pronounced in the 12-view setup, and furthermore, the relative gain compared to our initial reconstruction is the highest in perceptual quality (LPIPS).
Additionally, we report the results of our flow model applied to the test view poses (\methodname$++$) to highlight its effectiveness for refining renderings of unseen views.

\begin{figure*}[t]
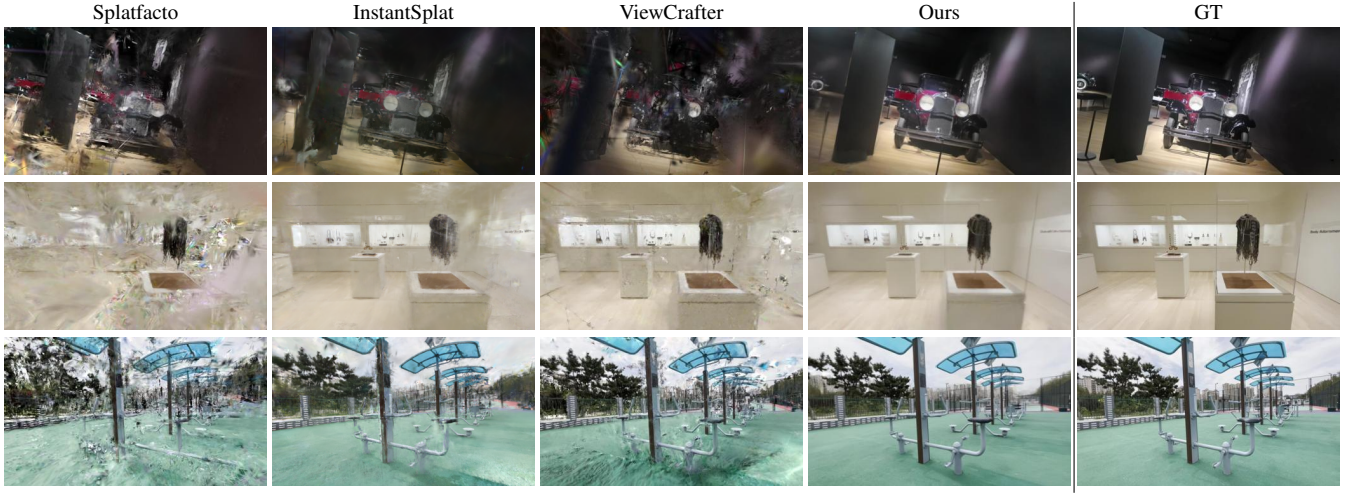

\centering
\footnotesize
\setlength{\tabcolsep}{1pt}
\begin{tabular}{@{}cccc|c @{}}
Splatfacto & InstantSplat & ViewCrafter & Ours & GT \\
\includegraphics[width=0.2\linewidth]{figures/qualitative/dl3dv/dac9796dd69e1c25277e29d7e30af4f21e3b575d62a0a208c2b3d1211e2d5d77/splatfacto/000018-rgb.jpg} &
\includegraphics[width=0.2\linewidth]{figures/qualitative/dl3dv/dac9796dd69e1c25277e29d7e30af4f21e3b575d62a0a208c2b3d1211e2d5d77/instantsplat/00018.png} &
\includegraphics[width=0.2\linewidth]{figures/qualitative/dl3dv/dac9796dd69e1c25277e29d7e30af4f21e3b575d62a0a208c2b3d1211e2d5d77/viewcrafter/000018-rgb.jpg} &
\includegraphics[width=0.2\linewidth]{figures/qualitative/dl3dv/dac9796dd69e1c25277e29d7e30af4f21e3b575d62a0a208c2b3d1211e2d5d77/ours/000018-rgb.jpg} &
\includegraphics[width=0.2\linewidth]{figures/qualitative/dl3dv/dac9796dd69e1c25277e29d7e30af4f21e3b575d62a0a208c2b3d1211e2d5d77/gt/000018-rgb_gt.jpg} \\

\includegraphics[width=0.2\linewidth]{figures/qualitative/dl3dv/6e11e7f4fea305c7c4658d2c1f8df29e6f299330860cf48ffbf1c5ff8b96c0a8/splatfacto/000028-rgb.jpg} &
\includegraphics[width=0.2\linewidth]{figures/qualitative/dl3dv/6e11e7f4fea305c7c4658d2c1f8df29e6f299330860cf48ffbf1c5ff8b96c0a8/instantsplat/00028.png} &
\includegraphics[width=0.2\linewidth]{figures/qualitative/dl3dv/6e11e7f4fea305c7c4658d2c1f8df29e6f299330860cf48ffbf1c5ff8b96c0a8/viewcrafter/000028-rgb.jpg} &
\includegraphics[width=0.2\linewidth]{figures/qualitative/dl3dv/6e11e7f4fea305c7c4658d2c1f8df29e6f299330860cf48ffbf1c5ff8b96c0a8/ours/000028-rgb.jpg} &
\includegraphics[width=0.2\linewidth]{figures/qualitative/dl3dv/6e11e7f4fea305c7c4658d2c1f8df29e6f299330860cf48ffbf1c5ff8b96c0a8/gt/000028-rgb_gt.jpg} \\

\includegraphics[width=0.2\linewidth]{figures/qualitative/dl3dv/7da3db99059a436adf537bc23309a6a4db9e011696d086f3a64037f7497d9df7/splatfacto/000018-rgb.jpg} &
\includegraphics[width=0.2\linewidth]{figures/qualitative/dl3dv/7da3db99059a436adf537bc23309a6a4db9e011696d086f3a64037f7497d9df7/instantsplat/00018.png} &
\includegraphics[width=0.2\linewidth]{figures/qualitative/dl3dv/7da3db99059a436adf537bc23309a6a4db9e011696d086f3a64037f7497d9df7/viewcrafter/000018-rgb.jpg} &
\includegraphics[width=0.2\linewidth]{figures/qualitative/dl3dv/7da3db99059a436adf537bc23309a6a4db9e011696d086f3a64037f7497d9df7/ours/000018-rgb.jpg} &
\includegraphics[width=0.2\linewidth]{figures/qualitative/dl3dv/7da3db99059a436adf537bc23309a6a4db9e011696d086f3a64037f7497d9df7/gt/000018-rgb_gt.jpg} \\
\end{tabular}
\caption{\textbf{Qualitative reconstruction results on DL3DV140~\cite{ling2024dl3dv}.} We show several examples of test view renderings from reconstructions obtained from our baselines and our method \emph{on the more challenging 12-view split}.}
\label{fig:dl3dv_qualitative}
\end{figure*}

\begin{table}[t]
\centering
\footnotesize
    \centering
    \setlength{\tabcolsep}{1pt}
    \begin{tabular}{lcccccc}
    \toprule
     & \multicolumn{3}{c}{\textbf{12-view}} 
     & \multicolumn{3}{c}{\textbf{24-view}} \\
     Method 
     & PSNR $\uparrow$ & SSIM $\uparrow$ & LPIPS $\downarrow$ 
     & PSNR $\uparrow$ & SSIM $\uparrow$ & LPIPS $\downarrow$ \\
    \midrule
    Splatfacto~\cite{tancik2023nerfstudio} & 16.71 &	0.528 &	0.478 &	22.17 &	0.738 &	0.309 \\
    InstantSplat$^\dagger$~\cite{fan2024instantsplat} & \rd20.47 & \rd	0.698 &	\nd 0.297 &	19.57 &	0.710 &	0.326 \\
    ViewCrafter$^*$~\cite{yu2024viewcrafter} & 19.19 &	0.638 &	0.375 &	\rd 21.95 &	\rd 0.734 & \rd	0.298 \\
    \methodname{} (Initial) & \nd 20.86 &	\nd 0.715 &	\rd 0.333 &	\nd 24.30 &	\nd 0.818 &	\nd 0.252 \\
    \methodname  &  \fs 22.43 &	\fs 0.766 & \fs	0.280 &	\fs 25.13 &	 \fs 0.836 &	\fs 0.212 \\ \midrule
    \methodname++ & 22.60 &	0.793 &	0.261 &	25.33 &	0.863 &	0.193 \\
    \bottomrule
    \end{tabular}
    \caption{\textbf{Sparse-view 3D reconstruction on DL3DV140~\cite{ling2024dl3dv}}. $^\dagger$ official code with GT poses, $^*$ official code with GT poses and our refined reconstruction method. Additionally, we report results of our generative model refining the test views (\methodname++).}
    \label{tab:dl3dv140}
\end{table}
\begin{table}[t]
\footnotesize
    \centering
    \setlength{\tabcolsep}{4pt}
    \begin{tabular}{lccc}
    \toprule
     Method 
     & PSNR $\uparrow$ & SSIM $\uparrow$ & LPIPS $\downarrow$ \\
    \midrule
    Splatfacto~\cite{tancik2023nerfstudio} & 22.41 &	0.843 &	0.352 \\
    GANeRF~\cite{roessle2023ganerf} & \nd 23.95 & 0.856 & \nd 0.306 \\
    \methodname{} (Initial) & \rd 23.84 & \nd 0.860 & \rd 0.331 \\
    \methodname  & \fs 24.11 & \fs 0.870 & \fs 0.303 \\ \midrule
    GANeRF w/ GAN & 24.01 & 0.860 & 0.291 \\
    \methodname++ & \textbf{24.90} &	\textbf{0.922} &	\textbf{0.250} \\
    \bottomrule
    \end{tabular}
    \caption{\textbf{Dense-view 3D reconstruction on ScanNet$++$ validation~\cite{yeshwanth2023scannet++}.} We use the official training and testing splits. In addition, we report results of our flow-matching model refining the test views (\methodname++), similar to~\cite{roessle2023ganerf} (w/ GAN).}
    \label{tab:scannetpp}
\end{table}

In \cref{tab:scannetpp}, we compare to state-of-the-art dense-view methods on the ScanNet$++$ benchmark. This setting is very challenging, as the initial reconstruction benefits from a large set of source images, and the test view poses are specifically chosen to represent out-of-distribution viewpoints~\cite{yeshwanth2023scannet++}. Despite these challenges, our method improves significantly over our initial reconstruction results, outperforming  GANeRF~\cite{roessle2023ganerf}. When applying post-hoc refinement of rendered views (bottom two rows), we also outperform GANeRF by a large margin, improving our initial results dramatically with a most significant improvement in LPIPS.
In \cref{fig:dl3dv_qualitative} and \cref{fig:scannetpp_qualitative}, we show qualitative comparisons on the aforementioned benchmarks. Our method exhibits strikingly fewer artifacts from floaters and poorly reconstructed geometry than the baseline methods.

Finally, in \cref{tab:nerfbusters}, we report results on Nerfbusters~\cite{warburg2023nerfbusters}. The benchmark consists of completely disjoint training and testing trajectories that cover distinct view angles and possibly scene content. Thus, the evaluation excludes areas unseen during training for NVS quality assessment and also reports coverage, \ie the percentage of 3D points seen in the training images reconstructed by a method, on the test trajectories. 
Our initial reconstruction method outperforms other reconstruction baselines like Nerfacto and Splatfacto~\cite{tancik2023nerfstudio} but lags behind Nerfbusters. Our full method outperforms Nerfbusters in PSNR and SSIM even without any postprocessing w.r.t. coverage, \ie maintaining the same degree of coverage as our initial reconstruction. When applying simple opacity thresholding to our renderings (see~\cref{sec:app_experiments}), we outperform Nerfbusters by a large margin on both NVS metrics and coverage.

\subsection{Ablation Studies}
\label{sec:ablations}

\begin{table}[t]
\footnotesize
\centering
    \begin{tabular}{lcccc}
    \toprule
     Method 
     & PSNR $\uparrow$ & SSIM $\uparrow$ & LPIPS $\downarrow$ & Coverage $\uparrow$ \\
    \midrule
    Splatfacto~\cite{tancik2023nerfstudio} & 16.17 & 0.529 & 0.375 & \rd 0.924 \\
    Nerfacto~\cite{tancik2023nerfstudio} & 17.00  & 0.527 & 0.380 & 0.896 \\
    Nerfbusters~\cite{warburg2023nerfbusters} & \rd 17.99 & \rd 0.606 & \nd 0.250 & 0.630 \\
    \methodname{} (Initial) &  17.02 & 0.567 & 0.365 & \fs 0.932 \\
    \methodname{} & \nd 18.31 & \nd 0.607 & \rd 0.337 & \fs 0.932 \\
    \methodname{}* & \fs 18.94 & \fs 0.780 & \fs 0.181 & 0.680 \\
    \bottomrule
    \end{tabular}
    \caption{\textbf{Dense-view 3D reconstruction on Nerfbusters~\cite{warburg2023nerfbusters}.} We report view quality and coverage, \ie how many pixels of the 3D points seen during training the method reconstructed, on the test trajectories. *using opacity thresholding.}
    \label{tab:nerfbusters}
\end{table}

\begin{table*}[t]
\centering
\footnotesize
\begin{tabular}{lccccc|ccc|cc}
\toprule
& MASt3R & Graph $\viewgraph$ & Re-triang. & ADC & Train steps
 & PSNR $\uparrow$ & SSIM $\uparrow$ & LPIPS $\downarrow$ & $|\Gauss|$ & Pairs \\
\midrule
1 & - & - & - & \checkmark & 30K & 19.44 &	0.633 &	0.394 &	317K &	- \\
2 & \checkmark & - & - & - & 5K & 21.72 &	0.739 &	\rd 0.294 &	2.01M &	342 \\
3 & \checkmark & \checkmark & - & - & 5K & 21.76 &	0.74 & 	\nd 0.293 &	2.01M &	129 \\
4 & \checkmark & \checkmark & \checkmark & - & 5K & \nd 22.23 &	\fs 0.768 &	0.322 &	186K &	129 \\
5 & \checkmark & \checkmark & \checkmark & \checkmark & 5K & \fs 22.58 &	\nd 0.766 &	\fs 0.292 &	244K &	129 \\
6 & \checkmark & \checkmark & \checkmark & \checkmark & 30K & \rd 22.21	& \rd 0.752 &	0.295 &	286K &	129\\
\bottomrule
\end{tabular}
\caption{\textbf{Initial reconstruction $\mathcal G_\text{src}$}. We compare 5 different variants (2-6) of our reconstruction pipeline to a naive Splatfacto~\cite{tancik2023nerfstudio} baseline, reporting the average scores across our DL3DV~\cite{ling2024dl3dv} view splits. We observe that dense initialization (2) improves results dramatically, but incurs a large computational burden. We show that our introduced co-visiblity graph $\viewgraph$ (3), the triangulation of 3D points (4), and the short schedule with ADC (4 and 5) consistently improve results while dramatically decreasing computational complexity. }
\label{tab:ablation_recon_stage1}
\end{table*}
\begin{table}[t]
\centering
\footnotesize
    \setlength{\tabcolsep}{2pt}
        \begin{tabular}{lc|ccc|ccc}
            \toprule
            & \multirow{2}{*}{$p_0$} & \multicolumn{3}{c|}{DL3DV140} & \multicolumn{3}{c}{ScanNet++} \\
            
            & & PSNR $\uparrow$ & SSIM $\uparrow$ & LPIPS $\downarrow$ & PSNR $\uparrow$ & SSIM $\uparrow$ & LPIPS $\downarrow$ \\
            \midrule
            \rowcolor{gray!25} 
            - & - & \textit{23.35} & \textit{0.782} & \textit{0.220} & \textit{24.18} & \textit{0.861} & \textit{0.278}  \\ \midrule
            1 & $\normal(\zero, \identity)$ & 23.50	& 0.769	& 0.200 & 24.09 &	0.873 &	0.208  \\ 
            2 & $p_0(\latent|\cond)$  & \textbf{24.15} &	\textbf{0.808} & 	\textbf{0.180} & \textbf{25.53} &	\textbf{0.921} &	\textbf{0.188} \\ 
            \bottomrule
            \end{tabular}
        \caption{
            \textbf{Different source distributions.} We report results on ScanNet++~\cite{yeshwanth2023scannet++} validation and the average of our DL3DV140~\cite{ling2024dl3dv} view splits at 512px resolution. In the first row, we show the scores of our initial reconstruction $\mathcal G_\text{src}$ which serves as input to our model as reference. Our conditional formulation improves significantly over a model with a standard Gaussian as source distribution.
        }
        \label{tab:ablation_video_model}
\end{table}

\begin{figure*}[t]
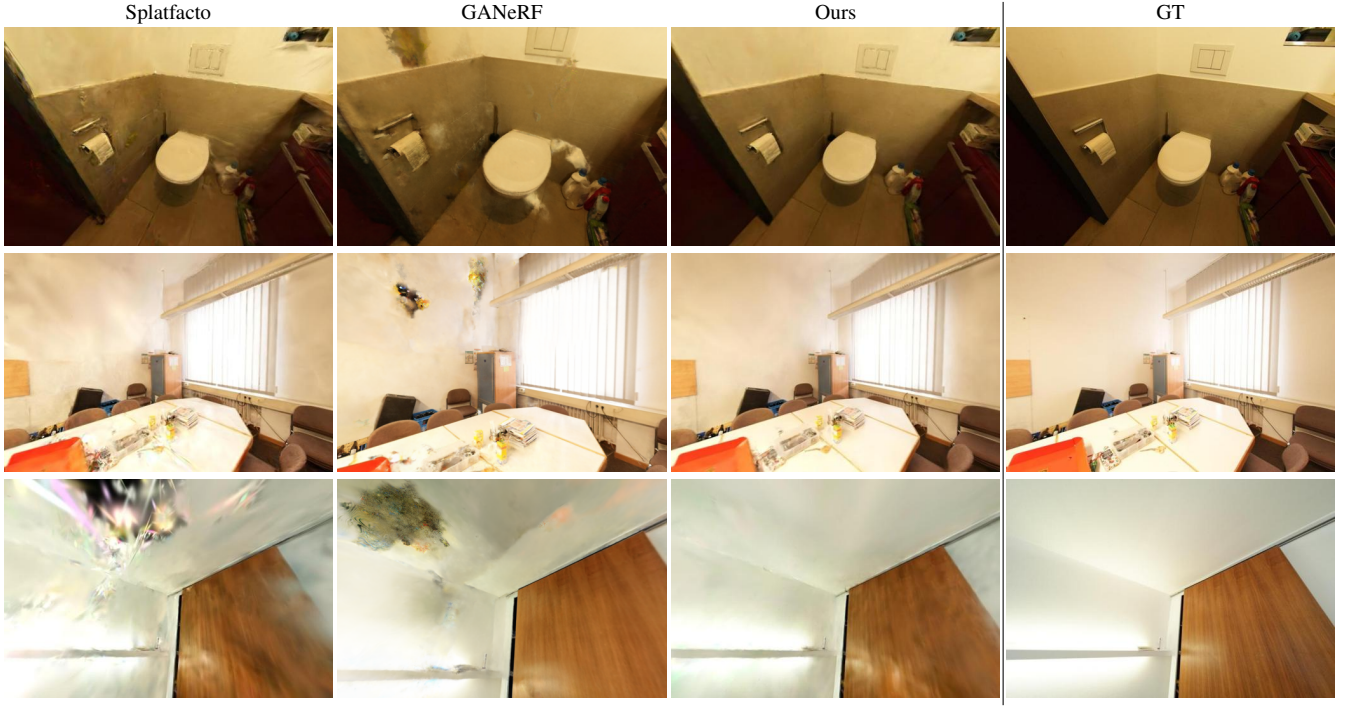

\centering
\footnotesize
\setlength{\tabcolsep}{1pt}
\begin{tabular}{@{}ccc|c @{}}
Splatfacto & GANeRF & Ours & GT \\
\includegraphics[width=0.25\linewidth]{figures/qualitative/scannetpp/286b55a2bf/splatfacto/000005-rgb.jpg} &
\includegraphics[width=0.25\linewidth]{figures/qualitative/scannetpp/286b55a2bf/ganerf/5.png} &
\includegraphics[width=0.25\linewidth]{figures/qualitative/scannetpp/286b55a2bf/ours/000005-rgb.jpg} &
\includegraphics[width=0.25\linewidth]{figures/qualitative/scannetpp/286b55a2bf/gt/000005-rgb_gt.jpg} \\

\includegraphics[width=0.25\linewidth]{figures/qualitative/scannetpp/f3d64c30f8/splatfacto/000001-rgb.jpg} &
\includegraphics[width=0.25\linewidth]{figures/qualitative/scannetpp/f3d64c30f8/ganerf/1.png} &
\includegraphics[width=0.25\linewidth]{figures/qualitative/scannetpp/f3d64c30f8/ours/000001-rgb.jpg} &
\includegraphics[width=0.25\linewidth]{figures/qualitative/scannetpp/f3d64c30f8/gt/000001-rgb_gt.jpg} \\

\includegraphics[width=0.25\linewidth]{figures/qualitative/scannetpp/25f3b7a318/splatfacto/000005-rgb.jpg} &
\includegraphics[width=0.25\linewidth]{figures/qualitative/scannetpp/25f3b7a318/ganerf/5.png} &
\includegraphics[width=0.25\linewidth]{figures/qualitative/scannetpp/25f3b7a318/ours/000005-rgb.jpg} &
\includegraphics[width=0.25\linewidth]{figures/qualitative/scannetpp/25f3b7a318/gt/000005-rgb_gt.jpg} \\
\end{tabular}
\caption{\textbf{Qualitative reconstruction results on ScanNet$++$~\cite{yeshwanth2023scannet++}.} We show \emph{out-of-distribution} test view renderings of the baselines and our method in the dense-view setting. This is a particularly challenging setting as the test views are far from the initial camera poses.   }
\label{fig:scannetpp_qualitative}
\end{figure*}

\parsection{Initial reconstruction $\mathcal G_\text{src}$.}
In \cref{tab:ablation_recon_stage1} we ablate the components of our reconstruction pipeline. Compared to the Splatfacto~\cite{tancik2023nerfstudio} baseline (1), using the dense 3D pointcloud from MASt3R~\cite{leroy2025grounding} yields sizable gains in performance across all metrics with LPIPS being improved particularly. However, there is a steep increase w.r.t. the number of Gaussians $|\Gauss|$. In addition, several hundreds of view pairs are needed to produce the initialization.
The co-visiblity graph $G_\text{vis}$ (3) decreases the number of required view pairs dramatically, even on the relatively small scale 12 and 24 view splits of DL3DV. 
Re-triangulating the matches (4) extracted from the MASt3R predictions, instead, dramatically reduces the number of Gaussians while increasing view quality in terms of PSNR and SSIM.
However, (4) exhibits a significantly higher LPIPS than variants (2) and (3). This is because while the triangulated 3D points are more precise, they exhibit holes in areas where geometry is uncertain. If we do not use ADC to grow Gaussians in these under-reconstructed areas, the LPIPS score will significantly increase.
Therefore, we enable ADC and achieve the best result in LPIPS, while performing similarly to (4) in terms of PSNR and SSIM. Finally, we show that our short schedule with 5k training steps is indeed sufficient to obtain a good reconstruction from an initial set of sparse views, since longer training (6) does not improve performance.

\parsection{Flow matching model.}
To show the importance of our formulation for view generation quality, we first implement a baseline model that retains a standard Gaussian distribution as $p_0$. In particular, we concatenate the input latents with the latents of the conditioning views, \ie target renderings and reference images, and the raymaps, following~\cite{blattmann2023stable}.
We compare the performance of this baseline to our model in~\cref{tab:ablation_video_model}. We show the NVS metrics of the initial renderings as reference in gray. While the baseline (1) improves significantly over the renderings in terms of LPIPS, the improvement is weaker in terms of PSNR and especially SSIM, where on DL3DV the model deteriorates the performance compared to the renderings.
Instead, our formulation (2) improves the renderings consistently across all metrics and outperforms the baseline by a significant margin.

\begin{table}[t]
\centering
\footnotesize
        \begin{tabular}{lcc|ccc}
        \toprule
        & $\loss_\text{tgt}$ & Re-init. & PSNR $\uparrow$ & SSIM $\uparrow$ & LPIPS $\downarrow$ \\
        \midrule
        1 & - & - & 23.53 &	0.795 &	0.279 \\
        2 & \checkmark & - & 23.72 &	0.799 &	0.248 \\
        3 & \checkmark & \checkmark & \textbf{23.78} &	\textbf{0.801} &	\textbf{0.246} \\
        \bottomrule
        \end{tabular}
        \caption{\textbf{Refined reconstruction $\mathcal G_\text{ref}$}. We report the average across our DL3DV~\cite{ling2024dl3dv} view splits. Both using $\loss_\text{tgt}$ and incorporating generated views in the pointcloud initialization (Re-init.) improve view synthesis quality.}
        \label{tab:ablation_recon_stage2}
\end{table}

\parsection{Refined reconstruction $\mathcal G_\text{ref}$.}
In ~\cref{tab:ablation_recon_stage2}, we test if differentiating the loss function between source and target images and using generated views for pointcloud initialization indeed improves reconstruction results.
To this end, we first train a model (1) on $\ims_\text{tgt}$ with the vanilla objective $\loss_\text{GS}$ and the pointcloud obtained from the source images. We then train a model (2) with our alternative loss function $\loss_\text{tgt}$ using the same initialization. Finally, we re-initialize the pointcloud (3) and train the model with it and $\loss_\text{tgt}$. The alternative loss applied to the generated target images significantly improves all NVS metrics with a particularly-pronounced improvement in LPIPS. Using the generated views for initialization leads to further improvement, however more pronounced in the 12 view setting (see \cref{sec:app_experiments}).

\section{Related Work}
We review works that approach 3D reconstruction and novel view rendering based on non-exhaustive scene captures.

\parsection{Geometric priors.}
To overcome the requirement of dense input views in~\cite{mildenhall2021nerf, kerbl20233d}, a line of works relied on geometry regularizations such as depth smoothness~\cite{niemeyer2022regnerf, kim2022infonerf}, ray entropy~\cite{barron2022mip, kim2022infonerf}, and re-projection losses~\cite{truong2023sparf,meuleman2023progressively} for a more well-behaved scene-level optimization. Other works use monocular depth and normals from off-the-shelf networks as geometry priors~\cite{roessle2022dense, yu2022monosdf, wang2023sparsenerf, zhu2024fsgs, meuleman2023progressively}. Recently, several works initialize 3DGS from either monocular~\cite{fu2024colmap} or multi-view networks ~\cite{fan2024instantsplat}. While our reconstruction pipeline aligns with these recent efforts, we propose a more scalable approach handling arbitrary view distributions and refine our reconstructions with generative priors.

\parsection{Feed-forward methods.}
Another line of work aims to directly predict novel views~\cite{yu2021pixelnerf, wang2021ibrnet, chen2021mvsnerf, hong2024lrm} from input images, skipping scene-level optimization entirely.
Along a similar axis, more recent works directly predict explicit 3D representations from input images~\cite{charatan2024pixelsplat, chen2024mvsplat, chen2024mvsplat360, tang2024lgm}. Instead of abandoning per-scene optimization, we propose a pipeline that integrates data priors both in the initialization and the training process of 3D Gaussian splatting.

\parsection{Generative priors.}
Generative priors enable improvements in both scene geometry and appearance. While earlier work used normalizing flows~\cite{niemeyer2022regnerf} and GANs~\cite{roessle2023ganerf}, many recent works focus on diffusion models~\cite{poole2023dreamfusion, wang2023prolificdreamer, wu2024reconfusion, gao2024cat3d, yu2024viewcrafter, chen2024mvsplat360, liu20253dgs}. In particular, many works incorporate diffusion models in the training of radiance fields~\cite{poole2023dreamfusion, wang2023prolificdreamer, wu2024reconfusion}, \eg via distillation~\cite{poole2023dreamfusion, wang2023prolificdreamer}. However, since this requires continuously evaluating the model during training, many recent approaches focus on directly generating additional training views~\cite{gao2024cat3d, yu2024viewcrafter, liu20253dgs, paul2024sp2360}. Among these, many works focus on multi-frame architectures since such models have shown great promise in video generation~\cite{blattmann2023stable, blattmann2023align, chen2024videocrafter2, yang2024cogvideox, menapace2024snap, polyak2025moviegencastmedia}, generating videos with high geometric consistency~\cite{li2024sora}. However, due to architectural limitations, these works only generate a handful of views at a time. Additionally, they keep the generative process as is and inject prior information into the model only as conditioning. We instead propose a flow matching process that directly refines initial reconstruction results and design the model architecture such that it still benefits from large-scale text-to-image pre-training.
\section{Conclusion}

We introduced \methodname, a novel pipeline that bridges the gap between sparse and dense 3D reconstructions. Our method enhances NVS by learning to map incorrect renderings to their corresponding ground-truth images. By training on a large-scale dataset of 3.6M image pairs, \methodname{} significantly improves 3DGS performance in both sparse- and dense-view scenarios, outperforming previous generative approaches that rely solely on 2D-conditioned synthesis.

\clearpage

\parsection{Acknowledgements.}
Yung-Hsu Yang was supported by the Swiss AI Initiative and a grant from the Swiss National Supercomputing Centre (CSCS) under project ID a03 on Alps.
The authors thank Andrew Brown, Corinne Stucker, and Duncan Zauss for helpful discussions and technical support. 

{\small
\bibliographystyle{ieee_fullname}
\bibliography{main}
}

\clearpage
\appendix
\section{Appendix}
\subsection{Data Details}
\label{sec:app_data}
To collect our large-scale dataset of image pairs, we utilize DL3DV10K~\cite{ling2024dl3dv} and ScanNet$++$~\cite{yeshwanth2023scannet++} benchmarks.
For DL3DV10K, we select $k \in [6, 36]$ equally spaced views as the initial sparse training set. We use the 960P resolution images and undistort them before reconstruction.
We filter sequences based on scale factor plausibility, \ie we remove sequences with very large or small scale factors. This makes our calibration pipeline robust to inaccuracies in initial camera parameters or depth estimates. We store the reconstructions in a compressed format following~\cite{morgenstern2024compact}. 
For ScanNet$++$, we first use farthest point sampling to determine a small number of keyframes and subsequently select 25\% - 50\% of the remaining training set closest to these keyframes. This ensures good spatial coverage while also promoting target views far from the initial training views, even in a dense view setting. Additionally, we use the out-of-distribution test views of the training sequences. We undistort the images and resize them to $640\times 960$. The resulting dataset covers a wide variety of input rendering quality for training, with initial PSNR values ranging from 5 to more than 30.

\subsection{Method Details}
\label{sec:app_method}

\parsection{Camera selection.}
For unordered view sets, we exploit the fact that our reconstructions are metric scale, which allows us to choose a reasonable range for the radius of the sphere that the candidate poses lie on. In practice, we found a range of $[0.2, 0.5]$ to work well, while for the orientation we use a random perturbation within $[0, 30]$ degrees in yaw and pitch.

\parsection{Initial reconstruction.}
Note that before running point triangulation, we check if there are sufficient reliable feature tracks inferred from the MASt3R matches, and if there are not enough feature tracks, we resort to global pointcloud alignment~\cite{leroy2025grounding} to ensure sufficiently dense 3D geometry estimates. This is usually only the case for very sparse input view scenarios, \ie 9 input views or less.

\begin{figure}[ht]
\centering
\includegraphics[width=\columnwidth]{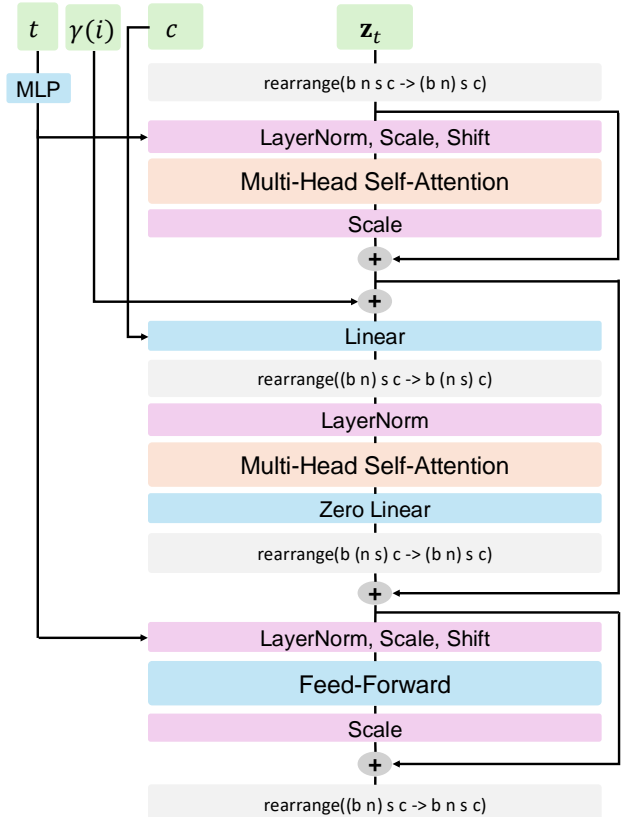}
\caption{\textbf{Illustration of our multi-view DiT block.} Here, we denote ray map embeddings as $c$.}
\label{fig:mv_dit_block}
\end{figure}

\parsection{Training details.}
For training our flow model, we found it important to choose the right camera selection strategy to ensure both sufficient viewpoint variability and high co-visibility between input views. Therefore, given a pool of target view candidates based on spatial similarity, we randomly sample the pool of candidates. Then, we select $k$ reference views based on a $k$-means clustering of the target views in a 6D space consisting of position and look-at direction. Given the $k$ clusters, we can choose the reference view closest to each cluster center. For timestep scheduling during training, we follow~\cite{esser2024scaling} and sample $t$ from a logit normal distribution with constant shift. Furthermore, we sample $t$ independently for each target frame~\cite{chen2025diffusion}, which we found to speed up convergence.

\begin{table}[t]
\centering
\footnotesize
    \begin{tabular}{ccccc}
        \toprule
        Num. views & GPU mem. (GB) &  PSNR $\uparrow$ & SSIM $\uparrow$ & LPIPS $\downarrow$ \\ \midrule
        72 & 42.3 & 24.15 & 0.808 & 0.180 \\
        12 & 17.0 & 24.29  & 0.801 & 0.175 \\
        \bottomrule
        \end{tabular}
    \caption{\textbf{Memory consumption vs. performance across input view counts.} We show our model is effective both at high and low number of views on DL3DV140 at 512px width (cf. \cref{tab:ablation_video_model}).}
    \label{tab:inference_memory}
\end{table}

\begin{table*}[t]
\centering
\footnotesize
    \begin{tabular}{lcc|cccccc}
    \toprule
    & \multirow{2}{*}{$\loss_\text{tgt}$} & \multirow{2}{*}{Re-init.} & \multicolumn{3}{c}{\textbf{12-view}} 
    & \multicolumn{3}{c}{\textbf{24-view}} \\
    & & & PSNR $\uparrow$ & SSIM $\uparrow$ & LPIPS $\downarrow$ & PSNR $\uparrow$ & SSIM $\uparrow$ & LPIPS $\downarrow$\\
     
    \midrule
    1 & - & - & 22.18 & 0.760 & 0.314 & 24.88 & 0.831 & 0.243 \\
    2 & \checkmark & - & 22.35 & 0.763 & 0.285 & 25.10 & 0.835 & \textbf{0.212} \\
    3 & \checkmark & \checkmark & \textbf{22.43} & \textbf{0.766} & \textbf{0.280} & \textbf{25.13} & \textbf{0.836} & \textbf{0.212} \\
    \bottomrule
    \end{tabular}
    \caption{\textbf{Refined reconstruction $\mathcal G_\text{tgt}$ ablation breakdown}. We report the scores on both DL3DV~\cite{ling2024dl3dv} view splits. Incorporating generated views in pointcloud initialization (Re-init.) benefits view synthesis, while the improvement is more pronounced in the 12 view setting.}
    \label{tab:ablation_recon_stage2_detail}
\end{table*}
\begin{table*}[t]
\centering
\footnotesize
\begin{tabular}{lccccccccc}
\toprule
 & \multicolumn{3}{c}{\textbf{3-view}} 
 & \multicolumn{3}{c}{\textbf{6-view}} 
 & \multicolumn{3}{c}{\textbf{9-view}} \\
 Method 
 & PSNR $\uparrow$ & SSIM $\uparrow$ & LPIPS $\downarrow$ 
 & PSNR $\uparrow$ & SSIM $\uparrow$ & LPIPS $\downarrow$ 
 & PSNR $\uparrow$ & SSIM $\uparrow$ & LPIPS $\downarrow$ \\
\midrule
ZipNeRF~\cite{barron2023zip}     
    & 12.77 & 0.271 & 0.705 
    & 13.61 & 0.284 & 0.663 
    & 14.30 & 0.312 & 0.633 \\ 
ZeroNVS~\cite{sargent2024zeronvs}    
    & 14.44 &  0.316 &  0.680 
    &  15.51 &  0.337 &  0.663 
    &  15.99 &  0.350 & 0.655 \\
ReconFusion~\cite{wu2024reconfusion}  
    & \nd 15.50 & \nd 0.358 & \nd 0.585 
    & \nd 16.93 & \rd 0.401 & \rd 0.544 
    & \nd 18.19 & \rd 0.432 & \rd 0.511 \\
CAT3D~\cite{gao2024cat3d}
    & \fs 16.62 & \fs 0.377 & \fs 0.515 
    & \fs 17.72 & \fs 0.425 & \fs 0.482 
    & \fs 18.67 & \fs 0.460 & \fs 0.460 \\
\methodname{}  & \rd 14.46 & \rd 0.347 & \rd 0.587 & \rd 16.18 & \nd 0.409 &  \nd 0.520 & \rd 17.53 & \nd 0.456 & \nd 0.467 \\ \midrule
\methodname{} (Initial) 
    & 12.77	& 0.243 &	0.592	
    & 14.40	& 0.320 &	0.532
    & 15.67	& 0.379 &	0.491 \\
\bottomrule
\end{tabular}
\caption{\textbf{Few-view 3D reconstruction on Mip-NeRF 360~\cite{barron2022mip}}. We follow the experimental setting of~\cite{wu2024reconfusion}.}
\label{tab:mipnerf}
\end{table*}

\parsection{Architecture details.}
The overall architecture of our model follows~\cite{esser2024scaling}. To compress the input images into latents, we use a VAE~\cite{rombach2022high} with latent dimension of 16. The latents are patchified with a patch size of 2 and fed to our latent flow matching model, which consists of 24 DiT blocks~\cite{peebles2023scalable}. Each block has 24 attention heads with dimension 64,  and the feedforward network has the same hidden dimension of 1536.
We show an illustration of our multi-view DiT block architecture \cref{fig:mv_dit_block}. As mentioned in ~\cref{sec:generate_reconstruct}, we keep the first self-attention layer, the normalization layers, and the feed-foward layer equivalent to~\cite{esser2024scaling}, and insert a multi-view attention layer after the first attention layer. This multi-view attention is concluded with a zero linear layer to keep the initialization intact. As mentioned in~\cref{sec:method_model}, for each source input view $i$, we condition the model on its camera pose $\pose_i$ and intrinsics $\intr_i$, \ie a pixel $\pix$ of image $i$ is represented as a ray $\mathbf{r} = \langle\mathbf{o} \times \mathbf{d}, \mathbf{d} \rangle$ with $\mathbf{o} = \rot_j^\top (\trans_i - \trans_j)$ and $\mathbf{d} = \rot_j^\top \rot_i \intr_i^{-1} \pix$ where $j$ is the reference view defining common coordinate frame.

\parsection{Inference details.}
When the number of reference views we can fit into a single forward pass is limited, we apply the same reference view selection strategy as in training. For target view selection, we use the method described in \cref{sec:generate_reconstruct} and use B-spline basis functions of degree 2. When inferring our model, we use 20 timesteps in the procedure described in \cref{sec:fm}, specifically:
\begin{equation}
    \latent_{t+\Delta t} = \latent_t + \Delta t \velocity_\theta(\latent_t, \mathbf{y}, t),
\label{eq:fm_inference}
\end{equation}
where the step size $\Delta t$ is chosen empirically as a monotonically decreasing function of $t$~\cite{esser2024scaling}.

\parsection{Runtime and memory analysis.}
As mentioned in \cref{sec:reconstruct_generate}, our initial reconstruction takes an average of 6.5 minutes for pointcloud initialization (6 minutes) and initial 3DGS training (30 seconds). Our multi-view flow model takes approximately 1.5 minutes to generate 200 additional images, processing 45 views at $540\times960$ resolution (91K tokens) on one H100 GPU at each forward pass. The final reconstruction training takes on average 42.4 minutes, as we use a longer, 30k step schedule and an additional LPIPS loss for the target views. In \cref{tab:inference_memory}, we report the memory usage of our model and show that it can run both on high-end and memory-constrained consumer grade GPUs by adjusting the number of input views that are processed in a single forward pass. Our model reaches comparable performance on our DL3DV140 benchmark with 72 and 12 views, respectively.

\parsection{Limitations.} While \methodname{} makes a significant step towards high-quality, photo-realistic 3D reconstructions from non-exhaustive captures, there remain meaningful directions for future work. For instance, our method relies on heuristics to select the camera views that are used to refine the 3D reconstruction results. In this regard, incorporating uncertainty quantification~\cite{jiang2024fisherrf, goli2024bayes} and active view selection~\cite{pan2022activenerf} could improve results. Additionally, because our method aims to map incorrect renderings to ground-truth images, its performance depends on the initial 3D reconstruction. In particular, if there are large areas entirely unseen in the source views, our model will not hallucinate new content. Incorporating a suitable prior for such cases opens up a promising avenue for future research. Finally, we focus on static scenes, as 3DGS exhibits major artifacts with dynamic objects. Hence, another meaningful direction for future work is to extend our method to dynamic scenes using a dynamic 3DGS method, which requires adapting the training data to reflect the presence of dynamic objects in both the source and target distributions.

\subsection{Additional Experiments}
\label{sec:app_experiments}

\parsection{Evaluation details.}
We use LPIPS with VGG-16 features unless otherwise specified in the benchmark, \ie we use VGG-16 for all experiments except for the Nerfbusters benchmark which uses AlexNet. Note that for Nerfbusters we resort to the test trajectory for target view selection since the test views are entirely disconnected from the initial reconstruction and as such it is not possible to effectively refine the reconstructed scene by interpolating along the training trajectory or by sampling poses around it. Finally, we optionally apply naive opacity thresholding, \ie we define a single minimum opacity value applied to all rendered views to achieve a comparable coverage to our baselines. The intuition behind this is that high opacity along a pixel ray usually correlates with well-defined scene geometry.

\parsection{Comparison to closed-source methods.}
In \cref{tab:mipnerf}, we compare with closed-source methods such as ReconFusion~\cite{wu2024reconfusion} and CAT3D~\cite{gao2024cat3d} using their provided data splits in MipNeRF360~\cite{barron2022mip}. For a fair comparison, we choose a similar camera selection strategy as CAT3D, where we generate an elliptic trajectory on a hemisphere around the common look-at point of the initial cameras. 
We note that the evaluation setting of~\cite{wu2024reconfusion} is distinct from ours since the training and evaluation splits are chosen so that there is a large fraction of the scene in the test views that was \emph{not observed in the training views}. As such, an evaluation with view synthesis metrics is only approximate, as there are many plausible 3D scenes for a set of partial observations. 
We show that our method, despite not being tailored for scene extrapolation as mentioned in~\cref{sec:app_method}, performs competitively to prior works. We further observe that the gap between our method and the state-of-the-art approach narrows when increasing the number of input views, where our method is almost on-par with CAT3D~\cite{gao2024cat3d} in terms of SSIM and LPIPS in the 9-view setting.

\parsection{Refined reconstruction ablation breakdown.}
As mentioned in \cref{sec:ablations}, we observed that the benefit of incorporating generated views into the pointcloud reconstruction is more pronounced in the 12-view setting, as can be seen in \cref{tab:ablation_recon_stage2_detail} where we provide a breakdown of the two splits. We attribute this to the fact that with an increasing number of co-visible views, there are enough reliable matches to triangulate a good initialization from the source input images and adding more views therefore ceases to improve results.

\immediate\closein\imgstream

\end{document}